\RequirePackage{fix-cm}
\documentclass[twocolumn]{svjour3}          
\smartqed  
\usepackage{graphicx}
\usepackage{latexsym}
\usepackage{url}
\usepackage{xcolor}
\definecolor{newcolor}{rgb}{.8,.349,.1}
\usepackage{float}
	\definecolor{blue(munsell)}{rgb}{0.0, 0.5, 0.69}
\usepackage{caption}
\usepackage{multirow}
\usepackage[font={color=blue(munsell),bf},figurename=Fig.,labelfont={it}]{caption}

\usepackage{enumitem}

 \usepackage{algorithm}
 \usepackage{algpseudocode}
 \usepackage{algcompatible}
\usepackage{hyperref}
\usepackage{amsmath}
\hypersetup{
     colorlinks = true,
     linkcolor = blue,
     anchorcolor = blue,
     citecolor = red,
     filecolor = blue,
     urlcolor = blue
     }

\begin{document}

\title{A Novel Hand Gesture Detection and Recognition system based on
ensemble-based Convolutional Neural Network 
}

\author{Abir  Sen         \and
        Tapas Kumar Mishra \and Ratnakar Dash 
}


\institute{ Abir Sen \at 
                Department of Computer Science and Engineering, National Institute of Technology, Rourkela 769008, India\\
				 \email{tarusince92@gmail.com; 518cs1007@nitrkl.ac.in} 
           \and
           Tapas Kumar Mishra \at
              Department of Computer Science and Engineering, National Institute of Technology, Rourkela 769008, India\\
				 \email{mishrat@nitrkl.ac.in} 
		    \and
		    Ratnakar Dash\at
              Department of Computer Science and Engineering, National Institute of Technology, Rourkela 769008, India\\
				 \email{ratnakar@nitrkl.ac.in} 
}


\maketitle

\begin{abstract}
 Nowadays, hand gesture recognition has become an alternative for human-machine interaction. It has covered a large area of applications like 3D game technology, sign language interpreting, VR (virtual reality) environment, and robotics. But detection of the hand portion has become a challenging task in computer vision and pattern recognition communities. Deep learning algorithm like convolutional neural network (CNN) architecture has become a very popular choice for classification tasks, but CNN architectures suffer from some problems like high variance during prediction, overfitting problem and also prediction errors. To overcome these problems, an ensemble of CNN-based approaches is presented in this paper. Firstly, the gesture portion is detected by using the background separation method based on binary thresholding. After that, the contour portion is extracted, and the hand region is segmented. Then, the images have been resized and fed into three individual CNN models to train them in parallel. In the last part, the output scores of CNN models are averaged to construct an optimal ensemble model for the final prediction. Two publicly available datasets (labeled as Dataset-1  and Dataset-2) containing infrared images and one self-constructed dataset have been used to validate the proposed system. Experimental results are compared with the existing state-of-the-art approaches, and it is observed that our proposed ensemble model outperforms other existing proposed methods.

\keywords{Deep Learning \and Hand Gesture Recognition  \and Hand Detection \and Contour Extraction\and Gesture Segmentation \and Ensemble Learning}
\end{abstract}

\section{Introduction}
\label{A Novel Hand Gesture Recognition system based on 2D-Discrete Wavelet transformation and ensemble-based Convolutional Neural Network}

From the past few years, gesture recognition has boosted human-computer interaction (HCI) to facilitate interaction between human beings and computer devices. Nowadays, virtual reality is appearing in people’s daily life, and undoubtedly it will be the mainstream of human-machine interaction in the future \cite{Chen}. The principal reason for this is the low-cost availability of high-performing sensors. Microsoft Kinect V2 provides infrared, depth image, RGB color, and skeleton information from the captured scenes. Intel Realsense camera provides information about depth, RGB and, skeleton data, whereas a Leap Motion controller device can capture the infrared and skeleton data.
Most of the works have been performed by using the Leap Motion controller to improve human-machine interaction. For example, in \cite{pititeeraphab2016robot}, the authors have proposed a method for controlling the robot arm using the Leap Motion controller. Most researchers perform hand gesture recognition in real-time by analyzing hand-skeleton information provided by the Leap Motion device. All works along these lines are two-staged: the first stage is to compute feature descriptors then use any classifier to recognize the hand gestures. In \cite{lu2016dynamic}, it is shown that skeleton-based nodes are used as feature descriptors, and then they are fed into Hidden Conditional Neural Field (HCNF) classifier for classifying hand gestures. In \cite{mantecon2016hand}, the authors have proposed a new approach to recognize gestures for infrared images collected by using Leap Motion device, which has three stages: (1) the system builds image descriptor without any segmentation phase; (2) the resultant descriptor is dimensionally reduced by using compressive sensing algorithm; and (3) the reduced feature vectors are fed into support vector machine (SVM) classifier for final classification. 
 In \cite{chuan2014american}, the authors have proposed the American sign language recognition system with Leap Motion device, where k-nearest neighbor and SVM algorithms are used to classify 26 English Alphabet letters.
In \cite{huang2011gabor}, they have used the Gabor feature extractor and SVM classifier for performing gesture recognition tasks after the segmentation phase. \\
In most cases, the stages of the hand gesture recognition system are bounded to gesture detection and classification. But most of the approaches need to extract the features before classifying them. With the rapid growth in deep learning and computer vision, the researchers have chosen the CNN model for both feature extraction and classification and have achieved better classification results. \\
In \cite{yingxin2016robust}, a hand gesture recognition system is proposed based on canny edge detection for data pre-processing and CNN architecture to extract the features automatically. In \cite{wang2017large}, they have proposed a continuous hand gesture recognition system which is having two modules (1) first is the segmentation phase to segment the gestures; (2) the second is a recognition module to recognize the gestures using CNN.
In \cite{li2019hand}, they have applied CNN architecture to classify hand gestures as feature extractors and then to optimize the classification process, and they have used SVM to improve the performance. In \cite{fang2019gesture}, they have proposed a hand gesture recognition system based on CNN, and to solve the overfitting problem, deep convolutional generative adversarial network (DCGAN) is applied. They have trained on fewer samples and achieved better results. 
In \cite{neethu2020efficient}, they have detected the hand region portion from the whole image. After segmentation of the fingertips from the gesture images, they are fed into CNN classifier to classify the gesture images, and their proposed methodology had achieved 96.2\% of recognition rate. In \cite{hu2020deep}, they have presented a CNN model-based hand gesture recognition system for unmanned aerial vehicles (UAV) flight controls after collecting skeletal data by using the Leap motion controller, and their proposed work has achieved a gesture recognition rate of 97\%. 
\\
But there are some limitations of using CNN architectures during classifying gestures due to high variance during predictions, which leads to overfitting in the model. So there is an efficient way to reduce the variance, is to train multiple CNN models and then averaging the outputs from those models. This method is called as the ensemble learning method, which can effectively minimize variance problem and can produce some good classification results.\\
In this paper, we have presented a hand gesture recognition based on ensemble-based CNN model. The whole proposed work has three stages: (1) hand detection by background separation; (2) contour portion extraction and hand region segmentation; (3) classification of the hand gesture images by training three individual CNN architectures in parallel and then averaging the output scores of these models to build final ensemble model.
\\
 The remainder of this paper is organized as follows. The details of our proposed methodology are described in Section  \ref{Section-2}. The experimental details, dataset description, experimental results, and comparison with other schemes are described in Section \ref{Section-3}. Finally conclusion part is presented in Section \ref{Section-4}.

\section{Method overview \label{Section-2}}

Our proposed method contains some important stages like gesture detection, binary thresholding, contour selection, hand portion extraction, resizing, fed the resized images into individual CNN models, evaluate the performance of three custom CNN architectures (GoogLeNet-like, AlexNet-like, VGGNet-like), then build an optimal ensemble model for classifying gesture images. The block diagram of our proposed framework is illustrated in Fig. \ref{block_diagram}. All the phases are discussed in this section.

%
%
\subsection{Image pre-processing}
\begin{figure*}[!ht]

\includegraphics[width=\textwidth,height=4cm]{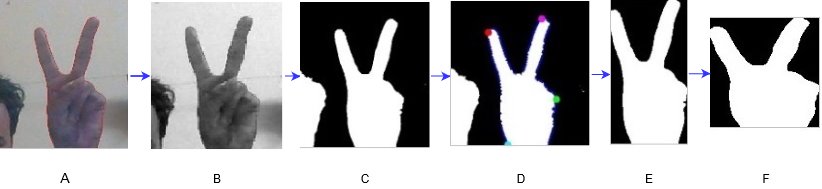}
\caption{\label{preprocessing} The whole pre-processing phase of generating hand region extraction before fed into CNN architecture (A) Gesture detection by background separation, (B) Transform to gray scale, (C) Binary thresholding, (D) Conrour region selection, (E) Hand portion extraction, (F) Resizing of images before applying median filter for noise removal.}
\end{figure*}

Image pre-processing is a significant phase to facilitate the model. This phase contains essential stages like gesture detection by background subtraction, binary thresholding, contour region extraction, hand portion segmentation, applying the median filter, and resizing of images before being fed into the CNN classifier. In Fig. \ref{preprocessing}, the whole pre-processing phase of generating resized images is shown. The different pre-processing steps are discussed below.
\vspace{-0.2cm}
\subsubsection{\label{detection}Gesture  detection by background subtraction}
\begin{figure}

\includegraphics[width=8.3cm,height=4cm]{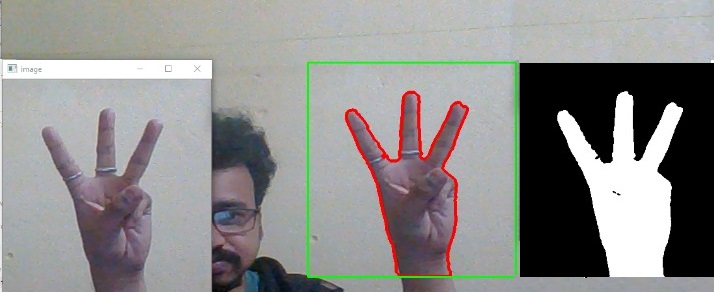}
\caption{\label{hand_detection} Real-time hand detection.}
\end{figure}

In image pre-processing, detection is a phase that identifies where the object is in the image and the boundaries. In our work, the main objective of gesture detection is to remove the unwanted background portion to extract the gesture portion. To detect the hand region, we have applied binary thresholding technique to segment between the background and foreground region, which makes the hand portion white in color and the background portion black in color. 
An example of real-time hand detection has been presented in In Fig. \ref{hand_detection}. 

\subsubsection{\label{Contour extraction}Contour portion and Hand region extraction}
\begin{figure}[!ht]
\includegraphics[width=8.3cm,height=4cm]{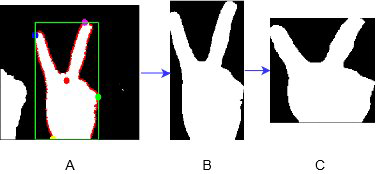}
\caption{ \label{fig:hand-gesture}Hand region extraction process (A) Contour region enclosed by bounded box, where red point represents palm center, (B) extraction of hand region portion, (C) resizing of images before fed into CNN classifier.}
\end{figure}

Contour region represents the boundary or outline of an object which is present in an image. The contour portion, having the largest area, is assumed to be the hand region.
The center of hand region (palm center) is estimated by using distance transform
\cite{chen2014real}, \cite {xu2017real} method.
According to this transform method, the pixel point in the contour region portion with having maximum value is considered as palm center. The palm radius is calculated by calculating the distance between the palm center and the point outside the contour region portion. In this way, the hand region is extracted from the arm portion. Fig. \ref{fig:hand-gesture} describes the whole procedure about contour portion selection and hand region segmentation. After extraction of the hand region portion, they are resized before fed into the CNN classifier for training.

\subsubsection{\label{resizing} Resizing}
Large images require more memory space and time than smaller images to be trained in the deep learning model. So in our work, after the gesture detection, contour extraction, and hand portion extraction phase, the images are resized into an aspect ratio of $(64 \times 64)$ before being fed into CNN models for both extracting features and classification.
\vspace{-0.2cm}

\begin{figure*}[!ht]
\centering
\includegraphics[width=14cm,height=7cm]{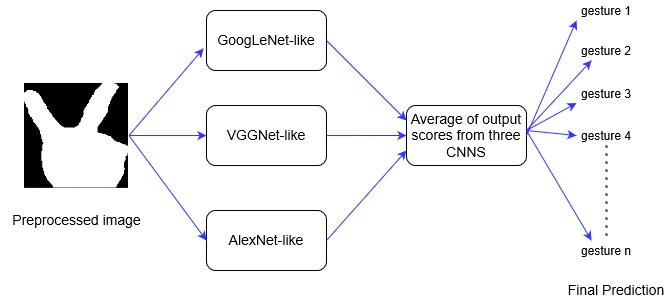}
\caption{\label{block_diagram} Block diagram of our proposed framework for gesture classification.}
\end{figure*}
\subsection{Ensemble model for classifying gestures}
We have designed an optimal ensemble based CNN model by averaging the output scores from three CNN architectures. We have constructed our self-designed CNN models based on GoogLeNet \cite{szegedy2015going}, VGGNet \cite{simonyan2014very}, and AlexNet \cite{krizhevsky2012imagenet} architectures. As we know that the CNN architectures like GoogLeNet, VGGNet provide powerful performance by taking lots of parameters for computing compared to other CNN architectures. In this section, we will discuss some self-designed CNN architectures: AlexNet-like, VGGNet-like, and GoogLeNet-like and then ensemble model.
\vspace{-0.2cm}
\paragraph{\textbf{AlexNet-like:}}
\begin{figure*}[!ht]

\includegraphics[width=\textwidth,height=4.5cm]{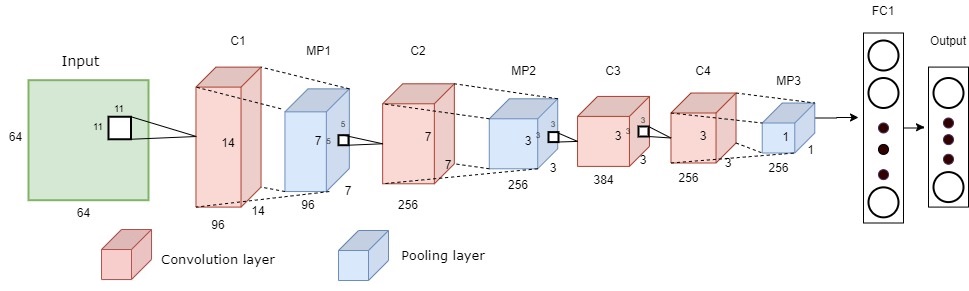}
\caption{ \label{alexnet-like} AlexNet-like architecture (One of the 3 self-constructed CNN architectures) used in our proposed technique.}
\end{figure*}

\begin{figure*}[!ht]

\includegraphics[width=\textwidth,height=5cm]{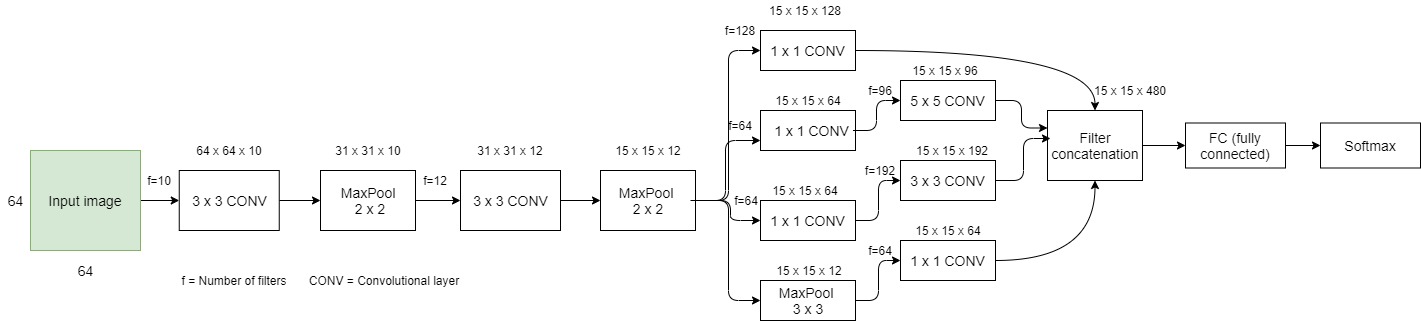}
\caption{ \label{googlenet-like}GoogLeNet-like architecture structure.}
\end{figure*}

 Classifying images from datasets like CIFAR10 and CIFAR100, is a simple task in the deep learning era. However, when there is the availability of a large dataset like ImageNet \cite{deng2009imagenet} consisting of millions of images, it will become a very challenging task. To train millions of labeled images, AlexNet \cite{krizhevsky2012imagenet}, a powerful CNN architecture, was proposed to extract the features by running the model on CUDA GPUs. It consists of eight layers, five are convolutional, and 3 are fully connected layers and have max-pooling layers followed by the Relu activation function.\\
 In our experiment, we have optimized AlexNet architecture into self-designed CNN architecture called AlexNet-like by reducing the input image size to ($64 \times 64 $) after the pre-processing phase. We have optimized the internal settings of the models, such as the number of filters, depth of architectures, which enhances the model performance by reducing the number of parameters. The model performance has been increased after reducing the number of parameters. The proposed CNN architecture is graphically illustrated in Fig. \ref{alexnet-like} and the model performance is shown in Table \ref{CNN1-compare}

\vspace{-0.2cm}
\paragraph{\textbf{VGGNet-like:}}
VGGNet architecture \cite{simonyan2014very} is a very powerful architecture that increases the depth of networks by adding convolution layers of filter size $(3\times3)$, max-pooling layers, batch-normalization layers, and dense layers and improves the model accuracy by adding more convolutional and dropout layers and it has achieved 92.7\% top-5 test accuracy in the ILSVRC-2012 classification challenge. This architecture can be used as transfer learning for large-scale datasets.
\\
Similarly, in our experiment, like AlexNet-like architecture, we have proposed self-designed VGGNet-like architecture, optimized by decreasing the depth of the network. After optimization, the number of parameters has been decreased, which makes the model easily trainable. The detailed configuration of our proposed VGGNet-like architecture for Dataset-1 and the number of trainable parameters at each layer are shown in Table \ref{vggnet-like}. The model performance is shown in Table \ref{CNN1-compare}.
In Table \ref{CNN1-compare}, we have shown the total number of parameters of VGG16 architecture on Dataset-1, Dataset-2, and self-constructed dataset.

\begin{table*}[!ht]
\caption{\label{vggnet-like} Detail configuration of our proposed VGGNet-Like architecture for Dataset-1.}
\begin{center}
 \resizebox*{\textwidth}{!}{
\begin{tabular}{cccccc}
\hline
layer type          & No of filters & Kernel size & Stride & Output shape            & No of parameters \\\hline \hline
Input image         & ---           & ---         & ---    & $64\times 64 \times 1$  & 0                \\
Convolution 2D      & 64            & $ 3\times 3 \times 64 $  & 1      & $64 \times 64 \times 64$       & 640              \\
Batch Normalization & ---           & ---         & ---    & $64\times 64 \times 64$       & 256              \\
Convolution 2D      & 64            & $ 3 \times 3 \times 64 $  & 1      & $64 \times 64\times 64$       & 36,928           \\
Batch Normalization & ---           & ---         & ---    & $64\times64\times 64$       & 256              \\
Max Pooling         & ---           & $2\times 2$       & 2      & $32\times 32\times 64$       & 0                \\
Convolution 2D      & 128           & $3\times 3\times 128$   & 1      & $32\times 32\times 128$      & 73,856           \\
Batch Normalization & ---           & ---         & ---    & $32\times 32\times 128$      & 512              \\
Convolution 2D      & 128           & $3\times 3\times 128$   & 1      & $32\times 32\times128$      & 147,584         \\
Batch Normalization & ---           & ---         & ---    & $32\times 32\times 128$      & 512              \\
Max Pooling         & ---           & $2\times2$       & 2      & $16\times 16\times 128$      & 0                \\
Convolution 2D      & 256           & $3\times 3\times 256$   & 1      & $16\times 16\times 256$      & 295,168         \\
Batch Normalization & ---           & ---         & ---    & $16\times 16\times 256$      & 1024             \\
Convolution 2D      & 256           & $3\times 3\times 256$   & 1      & $16\times 16\times 256$      & 590,080         \\
Batch Normalization & ---           & ---         & ---    & $16\times 16\times 256$      & 1024             \\
Convolution 2D      & 256           & $3\times 3\times 256$   & 1      & $16\times 16\times 256$      & 590,080         \\
Batch Normalization & ---           & ---         & ---    & $16\times 16\times 256$      & 1024             \\
Max Pooling         & ---           & $2\times2$       & 2      & $8\times 8\times 256$        & 0                \\
Convolution 2D      & 512           & $3\times 3\times 512$   & 1      & $8\times 8\times 512$        & 1,180,160        \\
Batch Normalization & ---           & ---         & ---    & $8\times 8\times 512$        & 2048             \\
Convolution 2D      & 512           & $3\times 3\times 512$   & 1      & $8\times 8\times 512$        & 2,359,808        \\
Batch Normalization & ---           & ---         & ---    & $8\times 8\times512$        & 2048             \\
Convolution 2D      & 512           & $3\times3\times 512$   & 1      & $8\times 8\times 512$        & 2,359,808        \\
Batch Normalization & ---           & ---         & ---    & $8\times 8\times 512$        & 2048             \\
Max Pooling         & ---           & $2\times2$       & 2      & $4\times 4\times 512$        & 0                \\
Flatten             & ---           & ---         & ---    & 8198             & 0                \\
Dense               & ---           & ---         & ---    & 512              & 4,194,816          \\
Dense               & ---           & ---         & ---    & 512              & 262,656           \\
Dense               & ---           & ---         & ---    & 10               & 5130             \\
Output              & ---           & ---         & ---    & ---              & 0                \\\hline
                    &               &             &        & Total parameters  &  12,107,466    \\\hline

\end{tabular}   
}
\end{center}

\end{table*}

\paragraph{\textbf{GoogLeNet-like:}}

GoogLeNet \cite{szegedy2015going} provides new inspiration to develop high-capability deep learning architecture.
It is a very powerful CNN architecture based on the inception module.  In case of inception module, the architecture is restricted by only $(1\times 1)$, $(3\times 3)$, $(5\times 5)$ filters. In this architecture, $(1\times 1)$ convolution filter is used before  $(3\times 3)$, $(5\times 5)$ convolutional filters. After that we have concatenated all $(1\times 1)$, $(3\times 3)$, $(5\times 5)$ 
convolutional filters to perform convolution on the output coming from the previous layer. Inception module also contains one max-pooling layer, then the outputs of all three filters are concatenated  (known as filter concatenation) and passed into the next layer as input.\\
In our experiment, we have built our custom CNN architecture called GoogLeNet-like architecture inspired by GoogLeNet. We have established an Inception architecture while optimizing the number of filters and layers. Hence, the model becomes easily trainable by reducing the number of parameters. The structure of this architecture is shown in Fig. \ref{googlenet-like} and the model performance is shown in Table \ref{CNN1-compare}.

\begin{figure}[!ht]
\centering
\includegraphics[width=8.3cm,height=6cm]{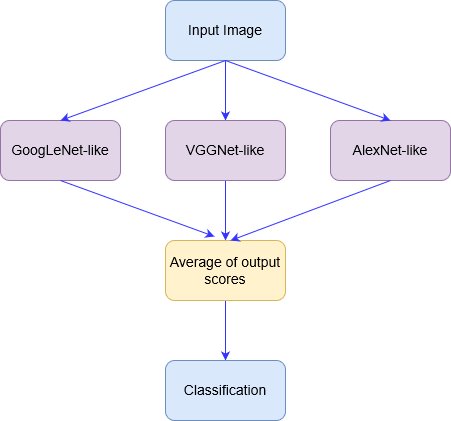}
\caption{Ensemble model.}
\label{ensemble}       
\end{figure}

\paragraph{\label{ECNN}\textbf{Ensemble model:}}
Ensemble learning is a popular machine learning technique that combines different learning models to reduce prediction errors and enhances model accuracy \cite{rajaraman2019performance}, \cite{polikar2012ensemble}. It is well known for being accurate and more robust than any individual CNN model.  In our experiment, we have used the parallel ensemble model as it has two types of methods, i.e., sequential and parallel. The technique uses three independent models in parallel, and then the model performance is improved by averaging the output scores of these three heterogeneous models. Then this average score is used to predict the final gesture label. The steps followed to build the ensemble model are listed in Algorithm \ref{algo:ensemble method}. Fig. \ref{ensemble} shows the flow diagram, how the output scores from three individual models (GoogLeNet-like, VGGNet-like, and AlexNet-like) are averaged, and to build final ensemble model for the classification task.
\begin{algorithm}
  	\caption{Ensembling method}\label{algo:ensemble method}
  	\begin{algorithmic}
  		\STATEx \textbf{INPUT}: Training data I=$(x_{i},y_{i})$, where i$\gets $ 1 to x, where x is number of training samples.\\
  		C is the number of classifiers.
  		 
  		\STATEx \textbf{OUTPUT}: Ensemble classifier E.
  		\STATE step 1: Learn the base classifiers.

  		\FOR{k $\gets$ 1 to C}
  		
  		\STATE Build the independent classifiers based on I.
  		\ENDFOR
  		\STATE step 2: Do the averaging of output scores from C number of models to learn the final classifier E.
  		
  		\STATE step 3: Utilize the classifier E for classification and final prediction.
  	\end{algorithmic}
  \end{algorithm}

\section{Experiment \label{Section-3}}

\subsection{Datasets}
\begin{figure}[!ht]
\includegraphics[width=8.3cm,height=4cm]{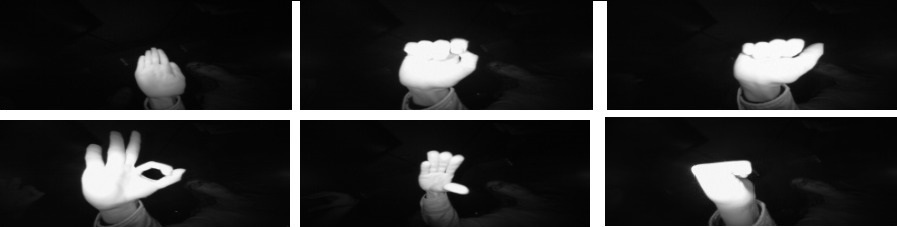}
\caption{\label{d1-sample} Hand gesture data samples.}
\end{figure}

Two publicly available gesture image datasets, hand gesture dataset \cite{mantecon2016hand} (labeled as Dataset-1), multi-modal hand gesture dataset \cite{mantecon2019real} (labeled as Dataset-2), and one self-constructed dataset have been used to evaluate our proposed model. Dataset-1 contains 20,000 images of 10 distinct gestures (Palm, L, Fist, Fist move, Thumb, Index, Ok, Palm move, Close, and Palm down) performed by ten different people (five men and five women). A total number of 200 images are recorded per gesture, So each person, there will be $200\times 10=2,000$ images. Here each infrared image has a resolution of $640 \times 240 $, in Fig. \ref{d1-sample}, we have shown some sample images from Dataset-1.

\begin{figure}[!ht]
\includegraphics[width=8.3cm,height=4cm]{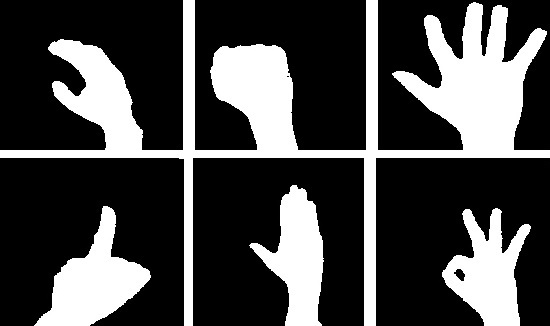}
\caption{\label{self-sample} Self-constructed dataset samples.}
\end{figure}

Dataset-2 contains 16 different gestures (Close, L, Palm down, Fist move, Five, Four, Hang, Heavy, Index, Ok, Palm, Palm move, Palm Up, Three, Two, and Up) performed by 25 people (17 men and eight women), where we have chosen only static gesture images after considering each static gesture has one kind of hand pose. Here all the gesture images have a non-identical resolution.
\\
In order to validate the proposed method, one self-constructed dataset containing 15,000 binary images captured by webcam has been used. This dataset contains 14 different gestures (Close, Fist, Five,  Four, Hang, Heavy, Index, L, Ok, Palm move, Three, Thumb, two, Palm) performed by three people (two men and one woman). Some self-constructed sample images are shown in Fig. \ref{self-sample}.
\vspace{-0.2cm}
\subsection{Evaluation protocol}
In our work, we have randomly split datasets into 80\% for training and 20\% for testing. Further, the training set is split into 75\% for training and 25\% for validation. Now in the dataset, 60\% is training images, 20\% is validation, and the remaining 20\% is testing images. The training image sets are used to fit the model and to study model parameters; validation image sets are used to estimate prediction errors during model selection, and testing images are used to evaluate the performance of the proposed ensemble model.
\vspace{-0.2cm}
\subsection{Experimental details}

For training our model, in the pre-processing phase, we perform binary thresholding to detect the gesture portion after removing the background section. After thresholding, we have extracted the contour regions in the resulting image, the region with the largest area represents the hand portion. Then median filter \cite{gupta2011algorithm} has been applied on the hand contour region to remove the noise. After that, the images are resized into a fixed aspect ratio of  $ 64 \times 64 $ and augmented before fed into three individual models (GoogLeNet-like, VGGNet-like, and AlexNet-like) for training. The augmentation includes rotations, zooming, horizontal and vertical-shift, and flipping during training of the model. While training each model, we have used softmax activation function, categorical cross entropy as loss function and Adam optimizer \cite{kingma2014adam} with an initial learning rate of 0.0001 and default values of $\beta_{1}$ =0.9, $\beta_{2}$=0.999. We have trained the training samples for 30 epochs with batch-size 160. Then the output scores of three CNN models are averaged to provide the final gesture label. Here we have used the model averaging ensemble method by using the “Average” class in Keras deep learning framework. The optimal values of parameters and hyper-parameters of three self-designed CNN models are listed in Table \ref{parameter_value} and the model diagram is shown in Fig. \ref{block_diagram}. All of our experiments have been performed on a system with Intel Core i7-10750H CPU, 16GB RAM having NVIDIA GeForce GTX 1650 GPU, and the whole program of our proposed model is developed under Keras deep learning framework.
\begin{table}[!ht]
\caption{\label{parameter_value} Summary of hyper-parameter settings.}
\begin{center}
    \begin{tabular}{|c|c|}
\hline
Parameter      & Value                     \\ \hline
Input size     & $64 \times 64$            \\ \hline
Epochs         & 30                        \\ \hline
Batch size     & 160                       \\ \hline
Learning rate  & 0.0001                    \\ \hline
Pooling     & $2\times 2$                       \\ \hline

Dropout     & 0.2                       \\ 
\hline

Optimizer      & Adam                      \\ \hline

Activation function      & ReLU                      \\ \hline

Error Function & Categorical cross entropy \\ \hline
\end{tabular}
\end{center}

\end{table}

\subsection{Transfer learning method}
Transfer learning is one of the most effective ways to classify large-scale datasets. It is considered to be fast and more straightforward than build a CNN architecture from scratch. In our work, we have considered some well-known pre-trained CNN architectures like VGG16 \cite{simonyan2014very}, AlexNet \cite{krizhevsky2012imagenet}, and GoogLeNet \cite{szegedy2015going} and then applied fine-tuning on them on the considered three datasets by using transfer learning. For this, we have considered the input shape of $(64\times64\times1)$ and the final FC (fully connected) layer of each CNN architecture have been replaced with a new FC (fully connected) layer containing $n$ nodes, where $n$ represents the number of classes of gesture images in Dataset-1, Dataset-2, and self-constructed dataset, respectively. In Table \ref{CNN1-compare}, these transfer learning based models are referred to as 
VGG16-TL, AlexNet-TL, and GoogLeNet-TL, respectively. In transfer learning, the CNN architectures are fine-tuned by using Adam optimizer with learning rate 0.0001 and default values of $\beta_{1}$=0.9, $\beta_{2}$=0.999. Here we have trained the samples for 30 epochs with batch-size 160.

\subsection{Results}

In this section, we have discussed the results from our every experiment over Dataset-1, Dataset-2, and self-constructed dataset by using our proposed model and then compared our model with several suitable methods in terms of confusion matrix, gesture recognition rate (\%), and accuracy. In our experiment, each column of the confusion matrix represents the actual gesture label, and the corresponding row represents the predicted gesture label.

\begin{figure*}[!ht]

\includegraphics[width=\textwidth,height=21cm]{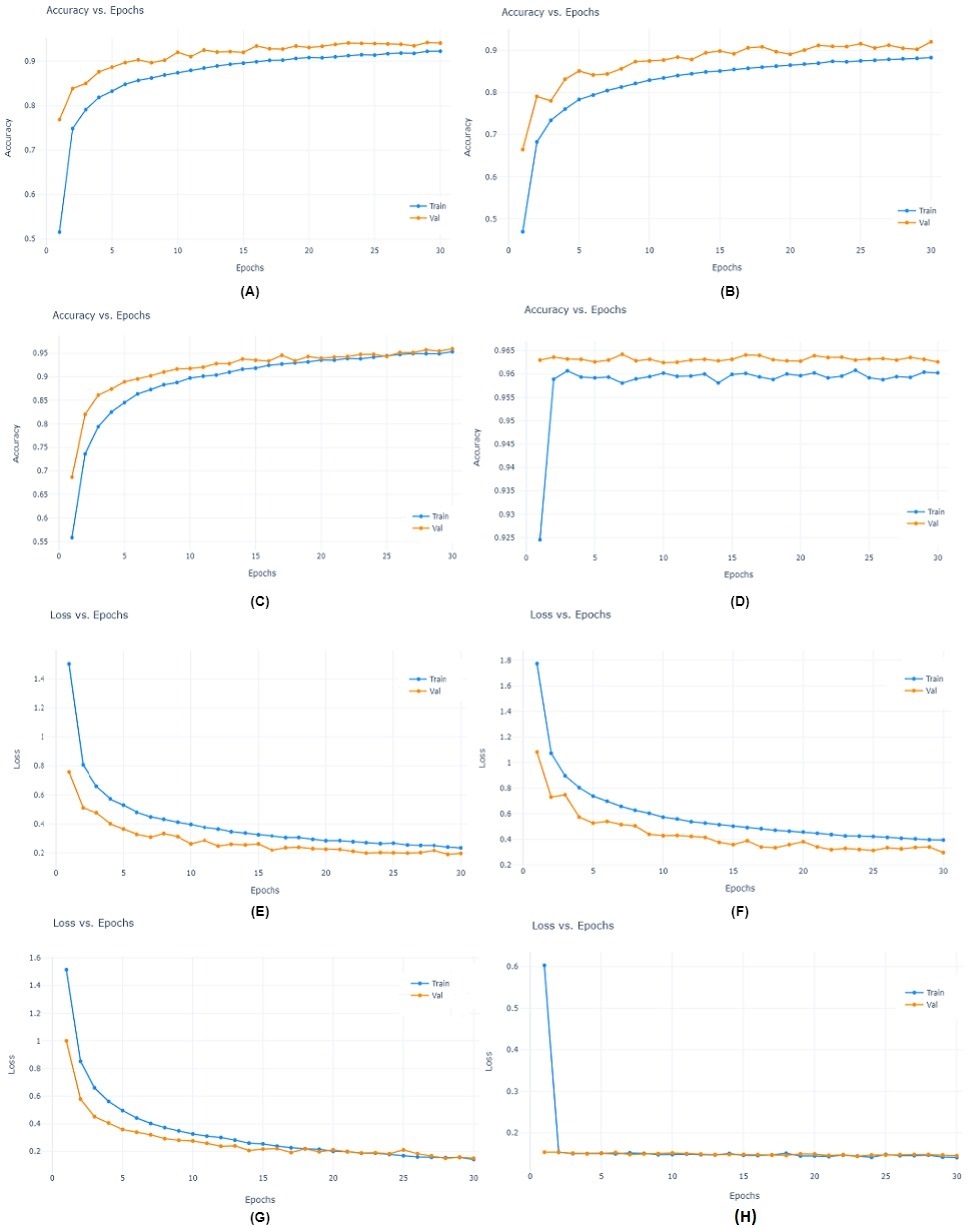}
\caption{\label{AC_curve1} Epoch-wise plots of accuracy and loss for individual models: Graphs A to D show the accuracy (training and validation) plots of the custom CNN models (i.e. AlexNet-like, GoogLeNet-like, VGGNet-like) and ensemble model, respectively. E to H show the loss occurs in case of each individual model and ensemble model for Dataset-2.}

\end{figure*}
Hand gesture classification accuracy is defined as: \\ \\
 Accuracy = $\frac{\text{Number of correctly classified gesture images}}{\text{Total number of testing samples}} \times 100 $ \\ \\
Table \ref{CNN1-compare} shows the testing results obtained by using
different CNN architectures over Dataset-1, Dataset-2, and self-constructed dataset. Each CNN model has been trained using different input size of $(32\times 32)$, $(64\times 64)$ and $(128 \times 128)$. It is
observed that the input image size of $(64\times 64)$ gives better classification results compared to other input image size.
 \\
The transfer learning results by using three pre-trained CNN models over three datasets have been listed in Table \ref{CNN1-compare}. It is observed that our proposed self-designed CNN architectures VGGNet-like, AlexNet-like, and \\ GoogLeNet-like architectures require less number of parameters and have faster training time (in sec) compared to transfer learning based models such as VGG16-TL, AlexNet-TL, and GoogLeNet-TL. \\
It is found in Table \ref{CNN1-compare} that our proposed ensemble model outperforms other CNN architectures and gives promising results in terms of gesture recognition accuracy(\%).
Fig. \ref{AC_curve1} shows the model accuracy and losses for individual models (i.e., AlexNet-like, GoogLeNet-like, VGGNet-like) and ensemble model respectively on Dataset-2 with respect to the number of epochs. In Fig. \ref{AC_curve1}(D), it is observed that the ensemble model has achieved an improved gesture recognition performance compared to individual CNN models.

\begin{figure}[!ht]
\includegraphics[width=8.3cm,height=6cm]{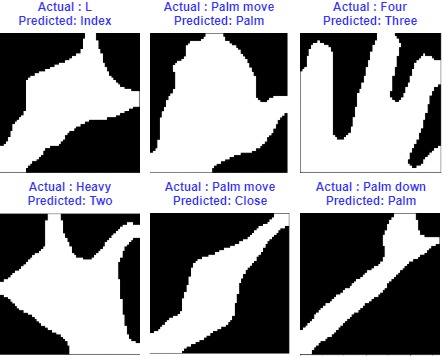}
\caption{\label{wrong_classiied}Visualization of some falsely classified images in Dataset-1, Dataset-2 and self-constructed dataset.}
\end{figure}

The analysis of the confusion matrix, shown in Tables \ref{confusion_1}, \ref{confusion_2} and \ref{confusion_3}, are carried out by using the proposed ensemble model. The analysis indicates some gestures, which produce misclassification errors. For instance,\\``Palm-move" and ``Palm" in Dataset-1 both are quite similar from viewing positions leading to a misclassification rate of 0.2\%. As an example, shown in Table \ref{confusion_1}, that ``Palm move" is misclassified as ``Palm" at a recognition rate of 0.2\%. The reason for this misclassification may occur due to the data augmentation technique making two gestures look similar. For another example shown in Table \ref{confusion_2}, that ``Fist move" is misclassified as ``Heavy" and ``Up" at a recognition rate of 0.5\% and 1\%. A similar situation also happens for gestures ``Five", ``Palm", ``Three", ``Two" and also ``Up". Similarly, in Table \ref{confusion_3}, it is observed that gesture ``Thumb" is misclassified as ``Five" and ``Four" at a recognition rate of 0.3\% and 1.1\%.
Some misclassified testing samples from Dataset-1, Dataset-2, and our self-constructed dataset are illustrated in Fig. \ref{wrong_classiied}.

\vspace{-0.2cm}

\begin{figure}[!ht]

\includegraphics[width=8.3cm,height=6cm]{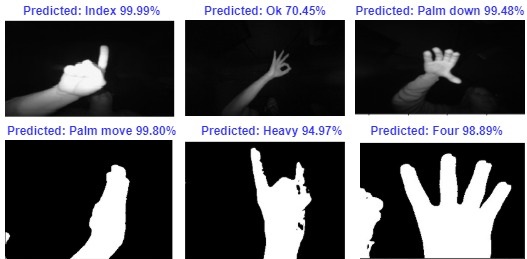}
\caption{\label{Truly_classiied} Performance of the proposed ensemble model on some testing samples from Dataset-1, Dataset-2 and our custom dataset.}

\end{figure}
                                                
\begin{table*}[!ht]
\caption{\label{CNN1-compare}Gesture recognition performance comparison results among different CNN models on Dataset-1, Dataset-2 and self-constructed dataset.}
\begin{center}
    \begin{tabular}{|c|c|c|c|c|c|c|c|c|c|}
\hline
\multirow{2}{*}{Model Structure}                                             & \multicolumn{3}{c|}{\begin{tabular}[c]{@{}c@{}}Dataset-1\\ (containing 10 gesture classes)\end{tabular}}                          & \multicolumn{3}{c|}{\begin{tabular}[c]{@{}c@{}}Dataset-2\\ (containing 16 gesture classes)\end{tabular}}                          & \multicolumn{3}{c|}{\begin{tabular}[c]{@{}c@{}}Self-constructed dataset\\ (containing 14 gesture classes)\end{tabular}}            \\ \cline{2-10} 
                                                                             & Parameters & \begin{tabular}[c]{@{}c@{}}Training\\ Time (s)\end{tabular} & \begin{tabular}[c]{@{}c@{}}Accuracy\\ (\%)\end{tabular} & Parameters & \begin{tabular}[c]{@{}c@{}}Training\\ Time (s)\end{tabular} & \begin{tabular}[c]{@{}c@{}}Accuracy\\ (\%)\end{tabular} & Parameters & \begin{tabular}[c]{@{}c@{}}Training\\ Time (s)\end{tabular} & \begin{tabular}[c]{@{}c@{}}Accuracy\\ (\%)\end{tabular} \\ \hline
Basic CNN                                                                    & 198,474    & 147                                                         & 94.72                                                   & 237,136    & 635                                                         & 82.79                                                   & 236,110    & 143                                                         & 83.10                                                   \\ \hline \hline
VGG16-TL                                                                     & 39,928,522 & 325                                                         & 99.65                                                   & 39,953,104 & 1120                                                        & 95.65                                                   & 39,944,910 & 280                                                         & 99.30                                                   \\ \hline
AlexNet-TL                                                                   & 28,810,690 & 220                                                         & 99.14                                                   & 28,816,696 & 840                                                         & 94.00                                                   & 28,814,694 & 150                                                         & 98.43                                                   \\ \hline
GoogLeNet-TL                                                                 & 7,223,540  & 280                                                         & 99.60                                                   & 7,236,146  & 900                                                         & 93.64                                                   & 7,231,944  & 180                                                         & 97.60                                                   \\ \hline \hline
VGGNet-like                                                                  & 12,107,466 & 273                                                         & 99.65                                                   & 12,110,544 & 1,020                                                       & 95.94                                                   & 12,109,518 & 240                                                         & 99.53                                                   \\ \hline
AlexNet-like                                                                 & 2,464,842  & 150                                                         & 99.36                                                   & 2,466,384  & 660                                                         & 94.21                                                   & 2,465,870  & 148                                                         & 98.53                                                   \\ \hline
GoogLeNet-like                                                               & 5,670,392  & 185                                                         & 99.50                                                   & 5,670,698  & 780                                                         & 93.38                                                   & 5,670,596  & 190                                                         & 97.20                                                   \\ \hline
\textbf{\begin{tabular}[c]{@{}c@{}}Ensemble model\\ (proposed)\end{tabular}} & 20,242,700 & 215                                                         & \textbf{99.80}                                          & 20,247,626 & 790                                                         & \textbf{96.50}                                          & 20,245,984 & 195                                                         & \textbf{99.70}                                          \\ \hline
\end{tabular}
    
\end{center}

\end{table*}



\begin{table*}[!ht]
\caption{\label{confusion_1}Confusion Matrix for gesture recognition rate (\%) using proposed ensemble model on Dataset-1.}

    \resizebox*{1.0\textwidth}{!} {
\begin{tabular}{|c|c|c|c|c|c|c|c|c|c|c|c|}
\hline
\multicolumn{1}{|l|}{}                                                       & \multicolumn{11}{c|}{Actual label}                                                                                                                                                \\ \hline
\multirow{11}{*}{\begin{tabular}[c]{@{}c@{}}Predicted \\ label\end{tabular}} &           & Palm           & L              & Fist           & Fist move      & Thumb          & Index          & OK            & Palm move     & Close          & Palm down      \\ \cline{2-12} 
                                                                             & Palm      & \textbf{100.0} & 0              & 0              & 0              & 0              & 0              & 0             & 0.2           & 0              & 0              \\ \cline{2-12} 
                                                                             & L         & 0              & \textbf{100.0} & 0              & 0              & 0              & 0              & 0             & 0             & 0              & 0              \\ \cline{2-12} 
                                                                             & Fist      & 0              & 0              & \textbf{100.0} & 0              & 0              & 0              & 0             & 0             & 0              & 0              \\ \cline{2-12} 
                                                                             & Fist move & 0              & 0              & 0              & \textbf{100.0} & 0              & 0              & 0             & 0             & 0              & 0              \\ \cline{2-12} 
                                                                             & Thumb     & 0              & 0              & 0              & 0              & \textbf{100.0} & 0              & 0             & 0             & 0              & 0              \\ \cline{2-12} 
                                                                             & Index     & 0              & 0              & 0              & 0              & 0              & \textbf{100.0} & 0             & 0             & 0              & 0              \\ \cline{2-12} 
                                                                             & Ok        & 0              & 0              & 0              & 0              & 0              & 0              & \textbf{98.6} & 0             & 0              & 0              \\ \cline{2-12} 
                                                                             & Palm move & 0              & 0              & 0              & 0              & 0              & 0              & 1.4           & \textbf{99.8} & 0              & 0              \\ \cline{2-12} 
                                                                             & Close     & 0              & 0              & 0              & 0              & 0              & 0              & 0             & 0             & \textbf{100.0} & 0              \\ \cline{2-12} 
                                                                             & Palm down & 0              & 0              & 0              & 0              & 0              & 0              & 0             & 0             & 0              & \textbf{100.0} \\ \hline
\end{tabular}
}
*Bold value represents the maximum recognition rate(\%) in each column

\end{table*}

\begin{table*}[]
\caption{\label{confusion_2}Confusion Matrix for gesture recognition rate (\%) using proposed ensemble model on Dataset-2.}
\resizebox*{1.0\textwidth}{!} {
\begin{tabular}{|c|c|c|c|c|c|c|c|c|c|c|c|c|c|c|c|c|c|}
\hline
                                  & \multicolumn{16}{c|}{Actual label}                                                                                                                                                                                                                                                                                                                                                                                                                           &               \\ \hline
\multirow{17}{*}{Predicted label} &                                                      & Close         & L             & \begin{tabular}[c]{@{}c@{}}Palm \\ down\end{tabular} & \begin{tabular}[c]{@{}c@{}}Fist\\ move\end{tabular} & Five          & Four          & Hang          & Heavy         & Index         & Ok            & Palm          & \begin{tabular}[c]{@{}c@{}}Palm\\ move\end{tabular} & \begin{tabular}[c]{@{}c@{}}Palm \\ Up\end{tabular} & Three         & Two           & Up            \\ \cline{2-18} 
                                  & Close                                                & \textbf{99.6} & 0             & 0                                                    & 0.1                                                 & 0             & 0             & 0             & 0             & 0             & 1.6           & 0             & 0                                                   & 0                                                  & 0             & 0             & 0             \\ \cline{2-18} 
                                  & L                                                    & 0             & \textbf{99.5} & 0.1                                                  & 0                                                   & 0             & 0             & 0.2           & 0             & 0.4           & 0             & 0             & 0                                                   & 0                                                  & 0.1           & 0.1           & 0             \\ \cline{2-18} 
                                  & \begin{tabular}[c]{@{}c@{}}Palm\\ down\end{tabular}  & 0             & 0             & \textbf{96.6}                                        & 0.1                                                 & 2.0           & 0             & 0.2           & 0             & 0             & 0             & 1.3           & 0                                                   & 0                                                  & 0             & 0             & 1.3           \\ \cline{2-18} 
                                  & \begin{tabular}[c]{@{}c@{}}Fist \\ move\end{tabular} & 0             & 0             & 0.1                                                  & \textbf{98.3}                                       & 0             & 0             & 0             & 0             & 0.1           & 0             & 0             & 0                                                   & 0.1                                                & 0             & 0             & 0.3           \\ \cline{2-18} 
                                  & Five                                                 & 0             & 0             & 0                                                    & 0                                                   & \textbf{96.6} & 0             & 0             & 0             & 0             & 0.1           & 0.1           & 0                                                   & 0.3                                                & 0             & 0             & 0             \\ \cline{2-18} 
                                  & Four                                                 & 0.1           & 0.2           & 0                                                    & 0                                                   & 0.7           & \textbf{99.9} & 0             & 0             & 0             & 0             & 0             & 0                                                   & 0                                                  & 4.1           & 0             & 0             \\ \cline{2-18} 
                                  & Hang                                                 & 0             & 0             & 0                                                    & 0                                                   & 0             & 0             & \textbf{99.5} & 0             & 0             & 0             & 0             & 0                                                   & 0                                                  & 0             & 0.2           & 0             \\ \cline{2-18} 
                                  & Heavy                                                & 0             & 0             & 0                                                    & 0.5                                                 & 0             & 0             & 0.1           & \textbf{99.8} & 0             & 0             & 0             & 0                                                   & 0                                                  & 0.1           & 0             & 0             \\ \cline{2-18} 
                                  & Index                                                & 0.1           & 0.2           & 0.2                                                  & 0                                                   & 0             & 0             & 0             & 0.1           & \textbf{99.3} & 0             & 0             & 0                                                   & 0                                                  & 0             & 0             & 0             \\ \cline{2-18} 
                                  & Ok                                                   & 0.1           & 0             & 0                                                    & 0                                                   & 0.5           & 0             & 0             & 0             & 0             & \textbf{96.0} & 0             & 0                                                   & 0                                                  & 0.2           & 0             & 0             \\ \cline{2-18} 
                                  & Palm                                                 & 0.1           & 0             & 1.4                                                  & 0                                                   & 0.7           & 0             & 0             & 0             & 0             & 0             & \textbf{97.1} & 2.0                                                 & 1.0                                                & 0             & 0             & 0             \\ \cline{2-18} 
                                  & \begin{tabular}[c]{@{}c@{}}Palm\\ move\end{tabular}  & 0             & 0             & 0.6                                                  & 0                                                   & 0             & 0.1           & 0             & 0             & 0             & 2.3           & 1.2           & \textbf{96.0}                                       & 2.8                                                & 0             & 0             & 0.2           \\ \cline{2-18} 
                                  & \begin{tabular}[c]{@{}c@{}}Palm \\ Up\end{tabular}   & 0             & 0             & 0                                                    & 0                                                   & 0             & 0             & 0             & 0.1           & 0             & 0             & 0.3           & 2.0                                                 & \textbf{94.8}                                      & 0             & 0             & 3.0           \\ \cline{2-18} 
                                  & Three                                                & 0             & 0.1           & 0                                                    & 0                                                   & 0             & 0             & 0             & 0             & 0             & 0             & 0             & 0                                                   & 0                                                  & \textbf{95.2} & 1.1           & 0             \\ \cline{2-18} 
                                  & Two                                                  & 0             & 0             & 0                                                    & 0                                                   & 0.1           & 0             & 0             & 0             & 0.2           & 0             & 0             & 0                                                   & 0                                                  & 0.3           & \textbf{98.6} & 0             \\ \cline{2-18} 
                                  & Up                                                   & 0             & 0             & 1.0                                                  & 1.0                                                 & 0             & 0             & 0             & 0             & 0             & 0             & 0             & 0                                                   & 1.0                                                & 0             & 0             & \textbf{95.2} \\ \hline
\end{tabular}
}
*Bold value represents the maximum recognition rate(\%) in each column
\end{table*}

\begin{table*}[!ht]
\caption{\label{confusion_3}Confusion Matrix for gesture recognition rate (\%) using proposed ensemble model on self-constructed dataset.}
\resizebox*{1.0\textwidth}{!} {
\begin{tabular}{|c|c|c|c|c|c|c|c|c|c|c|c|c|c|c|c|}
\hline
                                  &           & \multicolumn{14}{c|}{Actual label}                                                                                                                                                                                                                                          \\ \hline
\multirow{15}{*}{Predicted label} &           & Close          & Fist          & Five          & Four          & Hang           & Heavy          & Index          & L              & Ok             & \begin{tabular}[c]{@{}c@{}}Palm \\ move\end{tabular} & Three          & Thumb         & Two           & Palm          \\ \cline{2-16} 
                                  & Close     & \textbf{100.0} & 0             & 0             & 0             & 0              & 0              & 0              & 0              & 0              & 0                                                    & 0              & 0             & 0             & 0             \\ \cline{2-16} 
                                  & Fist      & 0              & \textbf{99.6} & 0             & 0             & 0              & 0              & 0              & 0              & 0              & 0.1                                                  & 0              & 0             & 0.4           & 0             \\ \cline{2-16} 
                                  & Five      & 0              & 0.4           & \textbf{98.9} & 2.1           & 0              & 0              & 0              & 0              & 0              & 0                                                    & 0              & 0.3           & 0             & 0             \\ \cline{2-16} 
                                  & Four      & 0              & 0             & 1.1           & \textbf{97.9} & 0              & 0              & 0              & 0              & 0              & 0                                                    & 0              & 1.1           & 0             & 0             \\ \cline{2-16} 
                                  & Hang      & 0              & 0             & 0             & 0             & \textbf{100.0} & 0              & 0              & 0              & 0              & 0                                                    & 0              & 0             & 0             & 0             \\ \cline{2-16} 
                                  & Heavy     & 0              & 0             & 0             & 0             & 0              & \textbf{100.0} & 0              & 0              & 0              & 0                                                    & 0              & 0             & 0             & 0             \\ \cline{2-16} 
                                  & Index     & 0              & 0             & 0             & 0             & 0              & 0              & \textbf{100.0} & 0              & 0              & 0                                                    & 0              & 0             & 0.4           & 0             \\ \cline{2-16} 
                                  & L         & 0              & 0             & 0             & 0             & 0              & 0              & 0              & \textbf{100.0} & 0              & 0                                                    & 0              & 0             & 0             & 0.4           \\ \cline{2-16} 
                                  & Ok        & 0              & 0             & 0             & 0             & 0              & 0              & 0              & 0              & \textbf{100.0} & 0                                                    & 0              & 0             & 0             & 0             \\ \cline{2-16} 
                                  & Palm move & 0              & 0             & 0             & 0             & 0              & 0              & 0              & 0              & 0              & \textbf{99.9}                                        & 0              & 0             & 0.4           & 0             \\ \cline{2-16} 
                                  & Three     & 0              & 0             & 0             & 0             & 0              & 0              & 0              & 0              & 0              & 0                                                    & \textbf{100.0} & 0             & 0             & 0             \\ \cline{2-16} 
                                  & Thumb     & 0              & 0             & 0             & 0             & 0              & 0              & 0              & 0              & 0              & 0                                                    & 0              & \textbf{98.6} & 0             & 1.1           \\ \cline{2-16} 
                                  & Two       & 0              & 0             & 0             & 0             & 0              & 0              & 0              & 0              & 0              & 0                                                    & 0              & 0             & \textbf{98.8} & 0             \\ \cline{2-16} 
                                  & Palm      & 0              & 0             & 0             & 0             & 0              & 0              & 0              & 0              & 0              & 0                                                    & 0              & 0             & 0             & \textbf{98.5} \\ \hline
\end{tabular}
}
*Bold value represents the maximum recognition rate(\%) in each column
\end{table*}

\begin{table*}[!ht]
\caption{ \label{comparison-1}
Gesture recognition performance comparison with other existing schemes on Dataset-1 in terms of accuracy results.}
\centering
\begin{tabular}{ccc}
\hline
Gesture Label & \begin{tabular}[c]{@{}c@{}}\textbf{Proposed}(\textbf{Background separation} + \textbf{hand portion extraction} +\\ \textbf{CNN} + \textbf{Ensemble method}) \end{tabular} & \begin{tabular}[c]{@{}c@{}}\cite{mantecon2016hand}(Feature descriptor\\ + SVM)\end{tabular} \\ \hline
Palm          & 1                                                                                                          & 1                                                               \\
L             & 1                                                                                                          & 0.990                                                                 \\
Fist          & 1                                                                                                          & 0.990                                                                 \\
Fist move     & 1                                                                                                          & 1                                                               \\
Thumb         & 1                                                                                                          & 0.990                                                                 \\
Index         & 1                                                                                                          & 1                                                                \\
Ok            & 0.990                                                                                                         & 0.990                                                             \\
Palm move     & 0.990                                                                                                           & 1                                                                \\
Close         & 1                                                                                                          & 1                                                                \\
Palm Down     & 1                                                                                                           & 1                                                                \\
Mean          & \textbf{0.998}                                                                                                         & 0.990               \\     \hline                                            
\end{tabular}
\end{table*}


\begin{table*}[!ht]
\caption{ \label{comparison-2}
Gesture recognition performance comparison with other existing proposed methods.}
\centering
\begin{tabular}{|c|c|c|}
\hline
Method                                                                   & Used datasets                                                                                                                                                            & Recognition rate \\ \hline
Feature descriptor + SVM \cite{mantecon2016hand}                                             & \begin{tabular}[c]{@{}c@{}}Leap motion sensor\\ captured infrared images \\ containing 10 gesture classes\end{tabular}                                                   & 99.0\%           \\ \hline

\begin{tabular}[c]{@{}c@{}}Segmentation + (HOG+ LBP)\\ +SVM \cite{mantecon2019real} \end{tabular} & \begin{tabular}[c]{@{}c@{}}Multimodal Leap motion\\ dataset (publicly available) \\ containing16 different hand poses\end{tabular}                                       & 96.0\%           \\ \hline
\multirow{3}{*}{\textbf{Proposed}}                                       & \begin{tabular}[c]{@{}c@{}}Leap motion sensor \\ captured infrared images\\ containing 10 gesture classes\end{tabular}                                                   & \textbf{99.8\%}  \\ \cline{2-3} 
                                                                         & \begin{tabular}[c]{@{}c@{}}Multimodal Leap motion\\ dataset (publicly available) \\ containing 16 different gesture classes\end{tabular}                                 & \textbf{96.4\%}  \\ \cline{2-3} 
                                                                         & \begin{tabular}[c]{@{}c@{}}Self-constructed dataset\\ where binary images collected by the \\ camera of computer,\\ containing 14 different gesture classes\end{tabular} & \textbf{99.7\%}  \\ \hline
\end{tabular}
\end{table*}

\vspace{-0.2cm}
\subsection{Comparison with other schemes}
To show the effectiveness our proposed model, we have compared our model with other existing schemes shown by \cite{mantecon2016hand} and \cite{mantecon2019real}  tabulated in Tables \ref{comparison-1} and \ref{comparison-2}. It is observed in Table \ref{comparison-1}, that our proposed work has achieved a better gesture recognition rate (in terms of accuracy results) than the technique proposed by Mantec{\'o}n T \cite{mantecon2016hand} over Dataset-1. 
In Table \ref{comparison-2}, it is observed that the proposed work by \cite{mantecon2016hand} used a combination of feature descriptor and SVM to perform classification of gestures at a recognition rate of 99.0\% and is evaluated on leap motion sensor captured infrared images containing ten gesture classes.

\par
The null hypothesis, in this case, is: the model is not statistically significant.
To check the statistical relevance in our experiment, we have performed one sample t-test by using IBM SPSS statistical analysis software. To get the value of  $t$, the following formula has been used:

\begin{center}
   $t = \frac{(\overline{X}-\mu)}{\frac{SD}{\sqrt{n}}}  $
\end{center}
where,
$\overline{X} $   mean of the samples \\
$\mu$  the test value \\
$SD$ sample standard deviation \\
$n$ sample size \\
\begin{table*}[!htbp]
\caption{ \label{one sample statistics}
One sample statistics.}
\begin{center}
\begin{tabular}{|l|l|l|l|}
\hline
\multicolumn{4}{|c|}{One sample statistics}  \\ \hline
N  & Mean   & Std deviation & Std Error Mean \\ \hline
10 & 99.8210 & 0.3088         & 0.0976          \\ \hline
\end{tabular}
\end{center}
\end{table*}
    
\begin{table*}[!htbp]
\caption{ \label{one sample test}
One sample test.}
\begin{center}
\begin{tabular}{|c|c|c|c|c|c|c|}
\hline
\multicolumn{7}{|c|}{One sample test}                                                                                                \\ \hline
      &    & \multicolumn{2}{c|}{Significance}   & Test value$=$99   & \multicolumn{2}{c|}{95\% Confidence interval of the difference} \\ \hline
t     & df & One-sided p      & Two-sided p      & Mean difference & Lower                          & Upper                          \\ \hline
8.405 & 9  & \textless{}0.001 & \textless{}0.001 & 0.82099         & 0.60003                        & 1.04196                        \\ \hline
\end{tabular}
\end{center}
\end{table*}

To calculate the value of $\overline{X} $, we have split Dataset-1 into ten parts and then evaluated the testing accuracy for each part, followed by calculating the mean (considered as sample mean) of these accuracy values with the ensemble model \ref{ECNN}.\\ As shown in Table \ref{one sample statistics}, the sample mean value ($\overline{X} $) is $99.821$, and the standard deviation (SD) is $0.3088$. The number of samples (sample size) is 10, and the test value ($\mu$) is 99. The results of the one-sample test are shown in Table \ref{one sample test}, where the test statistic of the one-sample t-test is denoted as $t$. Here $t$ = $8.405$, $t$ is calculated by using the above mentioned formula.\\
The results show that $p$ value $<$ 0.001, where $p$ value is used in hypothesis testing, to determine whether there is evidence to reject the null hypothesis by using calculated probability.
\par If $p$ $ < \alpha$ where, $\alpha$ (significance level) = 0.05, then null hypothesis is rejected. In Table \ref{one sample test}, it is observed that p-value is less than $\alpha$, where $\alpha$ = 0.05. So null hypothesis is rejected, and we can state that there is a statistically significant difference in the mean of the accuracy values.

The proposed work by \cite{mantecon2019real} (segmentation + (HOG + LBP) +SVM) has reported recognition rate of respectively 96.0\% after validating on publicly available multimodal Leap motion dataset. So our proposed work (background separation + hand region extraction + CNN + ensemble method) has achieved a promising average gesture recognition rate of 99.8\%, 96.4\%, and 99.7\% respectively after validating on Dataset-1 (leap motion sensor captured infrared images), Dataset-2 (multimodal Leap motion dataset), and self-constructed dataset compared to other existing schemes.
\vspace{-0.2cm}
\subsection{Our model performance on real-time video}
\begin{figure}[!htpb]
\centering
\includegraphics[width=5cm,height=21cm]{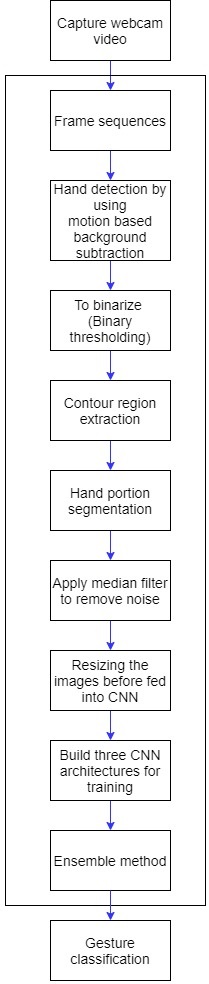}
\caption{\label{real-time_flow} Flow chart of real-time gesture recognition system.}
\end{figure}

\begin{figure*}[!ht]
\centering
\includegraphics[width=\textwidth,height=7.5cm]{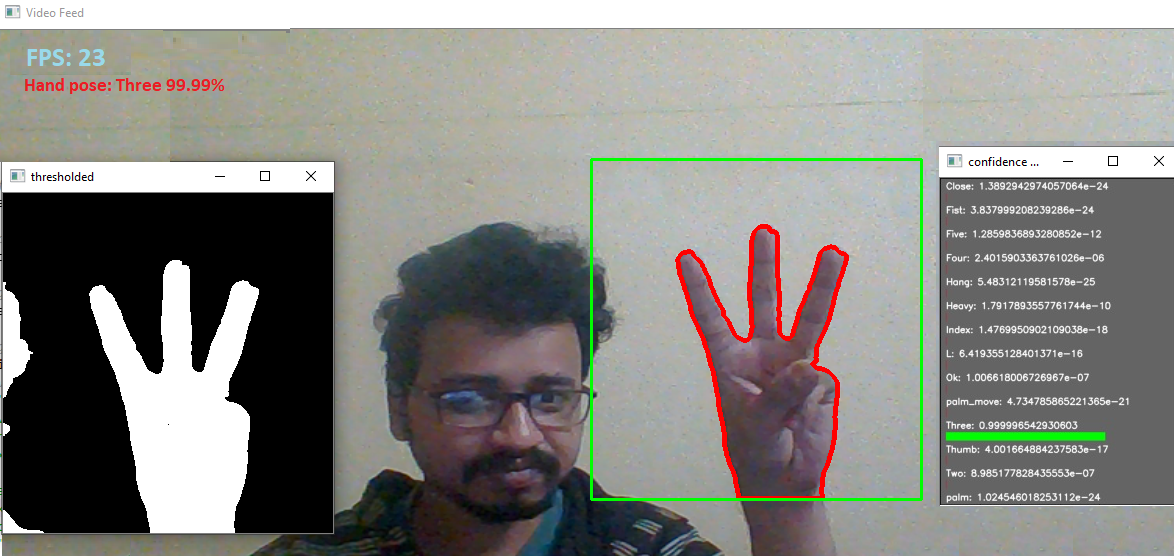}
\caption{\label{real-time} Gesture recognition performance of our proposed ensemble model on webcam video.}

\end{figure*}

To verify the effectiveness of our proposed model, we have deployed our trained model on the webcam-based video to perform real-time gesture detection and classification of gestures. This proposed model was implemented by using the system of having Intel Core i7-10750H CPU, 16GB RAM having NVIDIA GeForce GTX 1650 GPU. This work consists of six stages:\\
(1) Firstly, the method uses motion-based background separation technique \cite{rakibe2013background} in the video frame sequence to detect the hand portion after separating the unwanted background portion.\\
(2) To differentiate between foreground and background, we have applied binary thresholding, making the detected hand portion visible, whereas other unwanted parts remain black.\\
(3) After that, we have extracted the contour region from the detected image. The contour portion with having the largest area is considered to be the hand portion.\\
 (4) After hand portion segmentation, the median filter has been used to remove noise (shadow problem due to low light).\\
(5) Next, the images are resized into a fixed aspect ratio, and then they are fed into three individual CNN models for training.\\
 (6) In the last part, the output scores of the CNN models are averaged to build an ensemble model for classifying gesture images. The flow diagram of our real-time gesture recognition system is shown in Fig. \ref{real-time_flow} and the gesture recognition performance of our proposed method on real-time video is shown in Fig. \ref{real-time}. It is observed that our proposed ensemble model has been achieved a reasonable recognition rate in the presence of the low-light environment. The average recognition speed of our experiment has reached 20 fps while performing real-time gesture recognition by using the ensemble model.

\subsection{Real-time gesture recognition performance analysis}
Analysis of hand gesture recognition performance in the real-time scenario has been shown in Table \ref{real-time-1}. In this work, to show the average gesture classification time, we have taken three samples of each gesture, denoted as `Gesture sample 1', `Gesture sample 2' and `Gesture sample 3' for 5 seconds. Then we have calculated the delay between video capture and gesture classification results by the following formula: 
\begin{center}
Time delay = (time taken for video capture $-$ time taken for gesture classification) 
\end{center}
After calculating sample-wise time delay, we have calculated the average classification time for each gesture (in milliseconds).
Average gesture classification time
is calculated by the formula:
\begin{center}
Recognition time =  $\frac{\text{Average recognition time for each gesture}}{\text{Total number of gesture classes}} $ 
    
\end{center}
 For instance, three samples of gesture ``Ok'' have been taken for 5 seconds. The time delay between video capture and classification results for three gesture samples is 0.106 ms, 0.098 ms, and 0.091 ms, respectively. So the average gesture recognition time for gesture ``Ok'' is 0.099 ms. Similarly, for every gesture, we have shown that the processing time for average gesture classification is 0.117 ms by using the formula as mentioned above.
\begin{table*}[!ht]
\caption{ \label{real-time-1}
Real-time gesture recognition performance analysis along with average gesture classification duration.}
\centering
\begin{tabular}{|c|c|c|c|c|}
\hline
              & \multicolumn{3}{c|}{\begin{tabular}[c]{@{}c@{}}Average time delay between video capture\\  and \\ classification results in milliseconds (ms)\\ (duration for each sample collection is 5 sec)\end{tabular}} &                                                                                             \\ \hline
Gesture label & Gesture sample 1                                                   & Gesture sanple 2                                                   & Gesture sample 3                                                   & \begin{tabular}[c]{@{}c@{}}Average classification time for each gesture\\ (in ms)\end{tabular} \\ \hline
Ok            & 0.106                                                              & 0.098                                                              & 0.091                                                              & 0.099                                                                                       \\ \hline
Two           & 0.104                                                              & 0.165                                                              & 0.130                                                              & 0.133                                                                                       \\ \hline
Five          & 0.147                                                              & 0.136                                                              & 0.112                                                              & 0.131                                                                                       \\ \hline
Close         & 0.085                                                              & 0.078                                                              & 0.091                                                              & 0.084                                                                                       \\ \hline
Fist          & 0.125                                                              & 0.122                                                              & 0.120                                                              & 0.122                                                                                       \\ \hline
Hang          & 0.085                                                              & 0.092                                                              & 0.101                                                              & 0.093                                                                                       \\ \hline
Heavy         & 0.094                                                              & 0.101                                                              & 0.114                                                              & 0.103                                                                                       \\ \hline
L             & 0.123                                                              & 0.113                                                              & 0.207                                                              & 0.147                                                                                       \\ \hline
Index         & 0.131                                                              & 0.135                                                              & 0.177                                                              & 0.147                                                                                       \\ \hline
Three         & 0.115                                                              & 0.120                                                              & 0.130                                                              & 0.122                                                                                       \\ \hline
Four          & 0.103                                                              & 0.113                                                              & 0.107                                                              & 0.107                                                                                       \\ \hline
Palm move     & 0.140                                                              & 0.107                                                              & 0.097                                                              & 0.114                                                                                       \\ \hline
Thumb         & 0.096                                                              & 0.087                                                              & 0.091                                                              & 0.091                                                                                       \\ \hline
Palm          & 0.131                                                              & 0.128                                                              & 0.187                                                              & 0.148                                                                                       \\ \hline
\multicolumn{4}{|c|}{Average gesture classification time (in ms)}                                                                                                                                                           & 0.117                                                                                       \\ \hline
\end{tabular}
\end{table*}

\subsection{Discussions}
From our experiment, we have ensured that 
custom-CNN models can produce good results by hyper-parameter optimization, model structure, and using dropout in dense layers.
But CNN architectures suffer from high variance problem due to high dependence on training data and having overfitting problem: this leads to increase in bias and reduction in generalization. We have resolved this issue by training three self-designed CNN models (i.e., GoogLeNet-like, VGGNet-like, and AlexNet-like) in parallel and then averaging the output scores from CNN models to construct the ensemble model for classification tasks.\\
Table \ref{CNN1-compare} shows that our ensemble model has achieved a higher classification accuracy (99.80\% on Dataset-1, 96.50\% on Dataset-2, and 99.76\% on the self-constructed dataset) than other individual CNN models. Tables \ref{comparison-1} and \ref{comparison-2} also show the superiority of our proposed method by comparing the results with the state-of-the-art approaches for the classification of the gesture images.\\
To show the superiority of our proposed approach, we have deployed our trained ensemble model on the webcam-based video to perform a real-time gesture recognition system and has achieved a good recognition accuracy. The average gesture recognition speed has reached $20$ fps. The average recognition speed is slightly dropped through our ensemble method compared to individual CNN models because it is built by combining the output layers from three CNN models. But we have effectively improved the accuracy and stability of our proposed gesture recognition system. Recognizing gestures in a very complicated background is still a challenging task during the segment of the gesture portion. In the future, we will improve our model to tackle hand movements in a very complex environment.
Furthermore, our method can be extended by creating several combinations of models and selecting the best model combination to construct the ensemble model for the current task.

\section{\label{Section-4}Conclusion}
In this paper, an efficient gesture recognition system has been presented based on ensemble-based CNN. We have employed a set of stages to process the gesture image and segment the hand region before fed into three CNN classifiers for training in parallel. In the last part, the output scores of these three CNN models are averaged to build our ensemble model for final prediction. Two publicly available datasets (Dataset-1 and Dataset-2) and one self-constructed dataset have been used to validate the proposed approach. The results of our experiments exhibited the superiority of the ensemble-based CNN approach over existing schemes. Our proposed method has achieved 99.80\%, 96.50\%, and 99.76\% accuracy on Dataset-1, Dataset-2, and our custom dataset, respectively. The average recognition speed has reached $20$ fps, and the average gesture classification takes 0.117 ms, which meets the requirements of an efficient real-time gesture recognition system. Our proposed work can be beneficial to build HCI (human-computer interaction) based system, gesture-controlled home automation system, sign language translation (really helpful for people with hearing or speech impairment), and controlling robots. In our future work, we will extend our proposed framework to build a multi-modal fusion-based application by combining other modalities like speech, eye-gaze tracking, and face recognition.


%
%


\end{document}